\documentclass[submission,copyright,creativecommons]{eptcs}
\providecommand{\event}{ICLP 2023} 
\usepackage{amssymb}
\usepackage{iftex}

\usepackage{graphicx}
\usepackage{caption}
\usepackage{subcaption}
\ifpdf
  \usepackage{underscore}         
  \usepackage[T1]{fontenc}        
\else
  \usepackage{breakurl}           
\fi

\title{Towards One-Shot Learning for Text Classification using Inductive Logic Programming}
\author{Ghazal Afroozi Milani, Daniel Cyrus, Alireza Tamaddoni-Nezhad
\institute{Department of Computer Science, University of Surrey, United Kingdom}
\email{\{g.afroozimilani,d.cyrus,a.tamaddoni-nezhad\}@surrey.ac.uk}
}
\def\titlerunning{Towards One-Shot Learning for Text Classification using ILP}
\def\authorrunning{G. Afroozi Milani, D. Cyrus \& A. Tamaddoni-Nezhad}
\begin{document}
\maketitle

\begin{abstract}
With the ever-increasing potential of AI to perform personalised tasks, it is becoming essential to develop new machine learning techniques which are data-efficient and do not require hundreds or thousands of training data. In this paper, we explore an Inductive Logic Programming approach for one-shot text classification. In particular, we explore the framework of Meta-Interpretive Learning (MIL), along with using common-sense background knowledge extracted from ConceptNet. Results indicate that MIL can learn text classification rules from a small number of training examples, even one example. Moreover, the higher complexity of chosen example for one-shot learning, the higher accuracy of the outcome. Finally, we utilise two approaches, Background Knowledge Splitting and Average One-Shot Learning, to evaluate our model on a public News Category dataset. The outcomes validate MIL's superior performance to the Siamese net for one-shot learning from text.
\end{abstract}

\section{Introduction}
Machine learning, in particular, Deep Neural Networks (DNNs), has achieved outstanding outcomes in numerous real world applications such as image and text classification \cite{lecun2015deep,russell2022artificial}. However, these learning methods require a huge number of training instances, which is not always possible to be provided in advance. An automated software tool, for example, should be able to learn from a small number of interactions with the user in order to be efficiently customised with every new user's requirement.
Inductive Logic Programming (ILP) and, in particular, Meta Interpretive Learning (MIL) can learn human-readable hypotheses from a small amount of training data \cite{muggleton2018ultra,tamaddoni2021human}. This capability is very promising for medical and industrial usage, especially when we do not have access to a large amount of training data.
The main contribution of this study is to develop and evaluate a new learning algorithm, Meta-Goal Learner (MGL), based on the framework of Meta-Interpretive Learning (MIL), which can be utilised for one-shot learning from textual data along with using a common-sense knowledge-base (i.e. ConceptNet) as background knowledge for learning.

\section{Inductive Logic Programming}

Inductive Logic Programming (ILP) is a subfield of symbolic artificial intelligence, which works based on inductive reasoning to detect generalisations, rules, or models \cite{MUGGLETON94,nienhuys1997foundations}. In this method, the system learns from training examples with the help of background information and using logic programming \cite{tamaddoni2021human}. Indeed, the primary purpose of ILP \cite{MUGGLETON94}, like the other types of machine learning, is to induce a model/ hypothesis that can generalise training examples but, unlike them, uses logic programs to represent data and learns relations \cite{cropper2022inductive}. 
\paragraph{}Using logic programs brings several benefits to ILP. Firstly, as they represent hypotheses in ILP and are related to relational databases, they can support relational data like graphs. On the other hand, because of its expressivity, it can learn complex relational theories like cellular automata, event calculus theories, and Petri nets \cite{cropper2022inductive}. Secondly, it is similar to natural language; therefore, readable and valuable for explainable AI and scientific discovery (\cite{muggleton2018ultra}). 
\paragraph{}
    Every standard ILP setting consists of three input sets, including Background Knowledge (B), positive training examples ($E^+$), and negative training examples ($E^-$) \cite{Muggleton2009InverseEA,MUGGLETON94}. The output is a Hypothesis $H$ such that:

    \begin{itemize}
        \item It is necessary $\forall$  e $\in$  $E^+$  ${B \not\models}$ e
        \item ${H}$ is complete/ sufficient if $\forall$  e $\in$  $E^+$ and ${H \wedge B \models}$ e
        \item ${H}$ is strong consistent if $\forall$  e $\in$ $E^-$ and ${H \wedge B \not\models}$ e 
        \item ${H}$ is weak consistent if ${H \wedge B \not\models  \square}$
    \end{itemize}
    
\subsection{Meta-Interpretive Learning}
Meta-Interpretive Learning (MIL) \cite{MILHigherorderdyadic2015} is a type of Inductive Logic Programming which allows learning logic programs from background knowledge, training examples and a declarative bias called metarules. Metarules are datalog clauses with variables quantified over predicate symbols (i.e., second-order variables)  \cite{Patsantzis-Muggleton-2022,Morel2019TypedML}.\\ 

The advantage of MIL is its capability to (1) automatically introduce new sub-definitions when it learns a predicate definition \cite{Steve2017,MILHigherorderdyadic2015}. This is referred  to as Predicate Invention. (2) learning recursive clauses. MIL uses metarules to restrict hypotheses space. In other words, a meta-rule is a higher-order expression in which the form of permitted clauses in the hypothesis space is instructed\cite{Steve2017,muggleton2018meta}.\\

\label{def:Metarule}A \textbf{metarule} is a higher-order well formed formula as follows:

\begin{centering}
 $\exists\sigma  \forall\tau    P(s_1 , . . . , s_m ) \leftarrow . . . , Q_i (t_1 , . . . , t_n ), . . .$   \\
\end{centering}
where\space $\sigma , \tau$ are disjoint variables, $\rho$ and $\zeta$ are sets of predicate symbols and constants respectively $P, Q_i \in \sigma \cup \tau \cup \rho $ and $s_1, ..., s_m, t_1, ..., t_n \in \sigma \cup \tau \cup \zeta.$ Metarules are shown normally without quantifiers as: $P (s_1 , . . . , s_m ) \leftarrow . . . , Q_i (t_1 , . . . , t_n ), . . .$ . P and $Q_i$ are existentially quantified higher-order variables, while the other variables present universally quantified variables.\\

The standard MIL problem \cite{MILHigherorderdyadic2015} is defined as follows:
Given metarules (M), definite program background knowledge (B) and ground positive and negative unit examples ($E^+$, $E^-$),
\textbf{MIL} returns a higher-order datalog program hypothesis (H) if one exists such that:
$M\wedge B\wedge H \models E^+$ and
$M\wedge B\wedge H \not\models E^-$  is consistent.

\paragraph{}In ILP, selecting proper Background Knowledge (B) is essential for appropriate learning performance. However, this could be difficult or expensive as often this should be done by experts. There are two main problems related to having a balance background knowledge, including too little and too much Background Knowledge. If using too little Background Knowledge, we might lose the target hypothesis. Having too much irrelevant Background Knowledge is another ILP challenge that can increase hypothesis space size and decrease learning performance (\cite{cropper2020logical}).

\paragraph{}There are two methods to overcome this issue, one is to enable an ILP system to invent a new predicate instead of missing background knowledge \cite{stephen2015can}. The second method is to use transfer learning to discover knowledge that can help mitigate the effects of background knowledge shortage \cite{lin2014bias}. 

\section{Commonsense Knowledge and ConceptNet}
Commonsense knowledge is an ordinary information that helps people make sense of everyday situations without mentioning them in their communications. Due to its implicitness, capturing this knowledge is beneficial for designing an effective human-computer interface and various types of Artificial Intelligence (AI) \cite{ilievski2021dimensions, han2020semantic}.
These common-sense knowledge sources are in various forms and about different types of knowledge. Ilievski et al. \cite{ilievski2021dimensions} divided them into five categories, common-sense knowledge graphs ( e.g. ConceptNet and Atomic), common knowledge graphs and ontologies(  e.g. Wikidata and Yago ), lexical resources ( e.g. WordNet and Roget), visual common-sense sources ( e.g. Visual Genome), and Corpora and language models ( e.g. GenericsKB and Language models) as shown in Table \ref{commonsenseKnowledge}. 
\begin{table}[]
    \centering
    \begin{tabular}{|c|c|c|l|}
    \hline
         Category&            Source&       Relations& Example\\
         \hline
    Commonsense KGs&     ConceptNet&   37&        mother-related\_to-family\\
    &                    Atomic&       9&         Person X bakes bread-xEffect-eat food\\
    &                    Glucose&      10&        someone A -eats- something A\\
    &                    WebChild&     4groups&   eating -type of- consumption\\
    &                    Quasimodo&    78636&     pressure cooker-cook faster-food\\
    &                    SenticNet&    1&         cold food -polarity- negative\\
    &                    HasPartKB&    1&         dairy food -has part- vitamin\\
    &                    Probase&      1&         apple -is a- food\\
    &                    Isacore&      1&         snack food -is a- food\\
    \hline
    Common KGs&          Wikidata&     6.7k&      food -has quality- mouthfeel\\
    &                    YAGO4&        116&       banana chip -rdf:type- food\\
    &                    DOLCE&        1&         n/p\\
    &                    SUMO&         1614&      food -hyponym- food\_product\\
    \hline
    Lexical resources&   WordNet&      10&        food -hyponym- comfort food\\
    &                    Roget&        2&         dish -synonym- food\\
    &                    FrameNet&     8(f2f)&    eating -evoke- Ingestion\\
    &                    MetaNet&      14(f2f)&   food -has role- food\_consumer\\
    &                    VerbNet&      36(roles)& eating -haspatient- comestible\\
    \hline
    Visual sources&      Visual Genome& 42374&    food -on- plate\\
    &                    Flickr30k&    1&        eating place -confers with- their kitchen\\
    \hline
    Corpora\& LMs&       GenericsKB&    n/a&      Animals search for food\\
    &                    GPT-2&         n/a&      eating at home will not lead to weight gain\\
    &                    ChatGPT&       n/a&      Food provides nourishment: People eat food to..\\
    &                      &            &       (huge amount of information about eating)\\
    \hline
    \end{tabular}
    \caption{Overview of common-sense knowledge sources partly taken from \cite{ilievski2021dimensions}}
    \label{commonsenseKnowledge}
\end{table}
\paragraph{}ConceptNet is a multilingual common-sense knowledge graph derived from the crowd-sourced Open Mind Common Sense project and knowledge from other resources like WordNet, DBPedia, Games with a purpose, Wiktionary, OpenCyc and JMDict, connecting words and terms of natural language as nodes with labelled weighted edges with 37 types of relations \cite{ilievski2021dimensions}.\\

The standard definition of a Common-sense Knowledge Graph is as follows \cite{wang2021inductive}:\\
G= (V,E,R) is a Common-sense Knowledge Graph, where V, E and R are the node, edge and relation sets. Edges comprise triplets (h, r, t) where h and t are head and tail entities connected by relation $r: E = {(h,r,t)\mid h \in V,t \in V,r \in R}$, and nodes are along with a free-text description.\\

\begin{figure}[t]
    \centering  
    \includegraphics[width=0.7\linewidth]{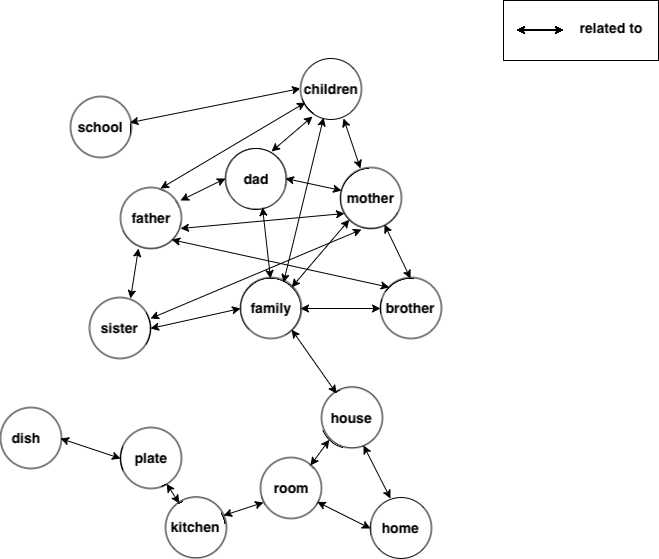}
    \caption{An illustration of ConceptNet used by our system. }
    \label{conceptnetexample}
\end{figure}

These node relations in ConceptNet5.5 have been categorised \cite{speer2017conceptnet} to two sub-classes: Symmetric relations like LocatedNear, RelatedTo, SimilarTo, and Asymmetric relations such as AtLocation, CapableOf, CreatedBy, DefinedAs, HasA, and Entails \cite{ilievski2021dimensions}. We use relations of "RelatedTo", which demonstrate an undefined meaning relation between two words. As shown in Figure \ref{conceptnetexample}, every node represents a word extracted from ConceptNet and connected to another node by a bidirectional edge. These symmetric edges show the relatedness of the connected words and, therefore, are bidirectional.

\section{One-shot text classification using ILP and MIL}

The basis of an interactive learning system is to learn from the user. This challenge requires incremental learning techniques that help the system to learn from time to time. In this section, we introduce Meta-Goal Learner (MGL), a novel machine learning from user interactions. We explore Metagol\cite{metagol} learning engine for the task of few-shot learning in MGL.\\

MGL initially fetches the words of the task (i.e. sentence given by the user) and obtains relevant background knowledge from ConceptNet. The system can then learn and categorise the upcoming tasks, as shown in Figure \ref{MGL}. For instance, when a user writes a task (e.g. "call mother"), the system will wait for the first answer/ label from the user and then will consider it as a positive example. In this case, the positive example represents the task and its category name. The system then generates relevant background knowledge as follows:
\begin{figure}[t]
    \centering  
     \begin{subfigure}[b]{0.45\textwidth}
    \centering
    \includegraphics[width=\linewidth]{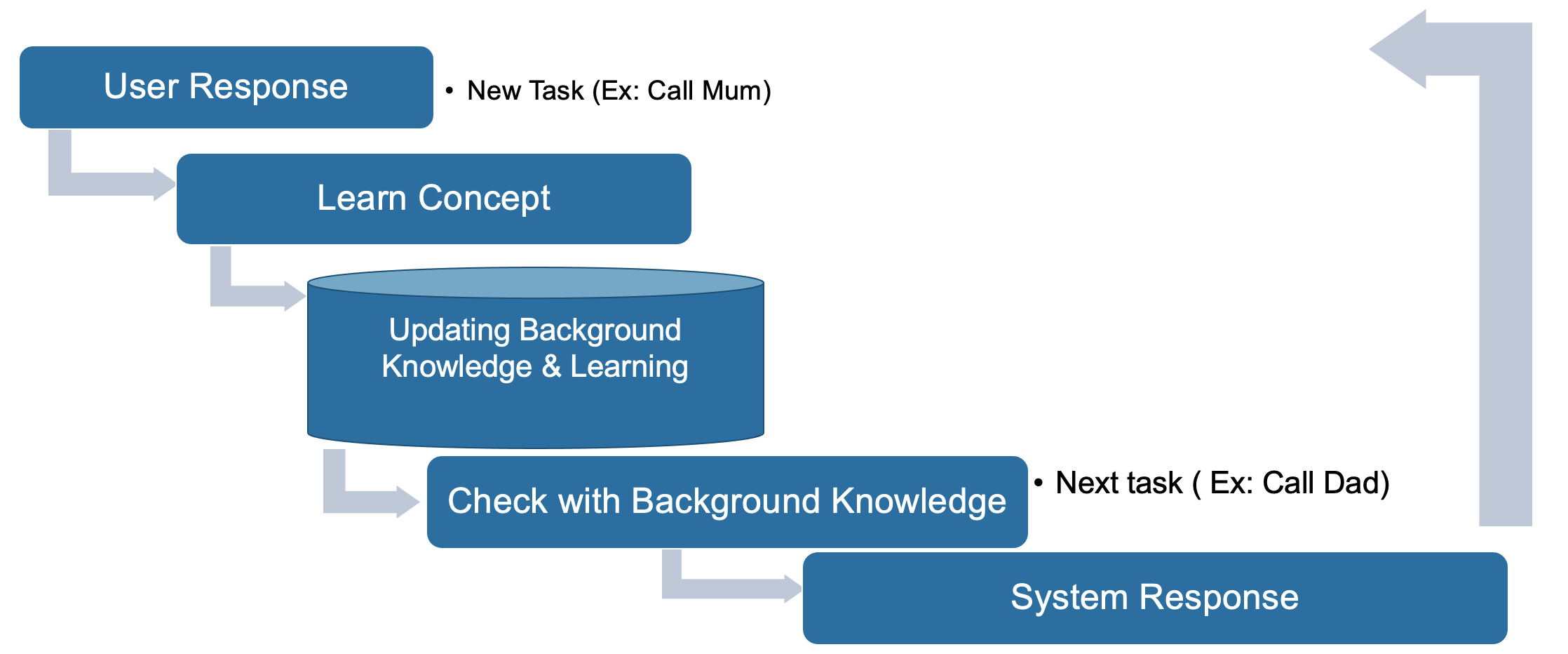}
     \caption{Model Prototype}
         \label{fig:nneg}
    \end{subfigure}
      \begin{subfigure}[b]{0.45\textwidth}
    \centering
    \includegraphics[width=\linewidth]{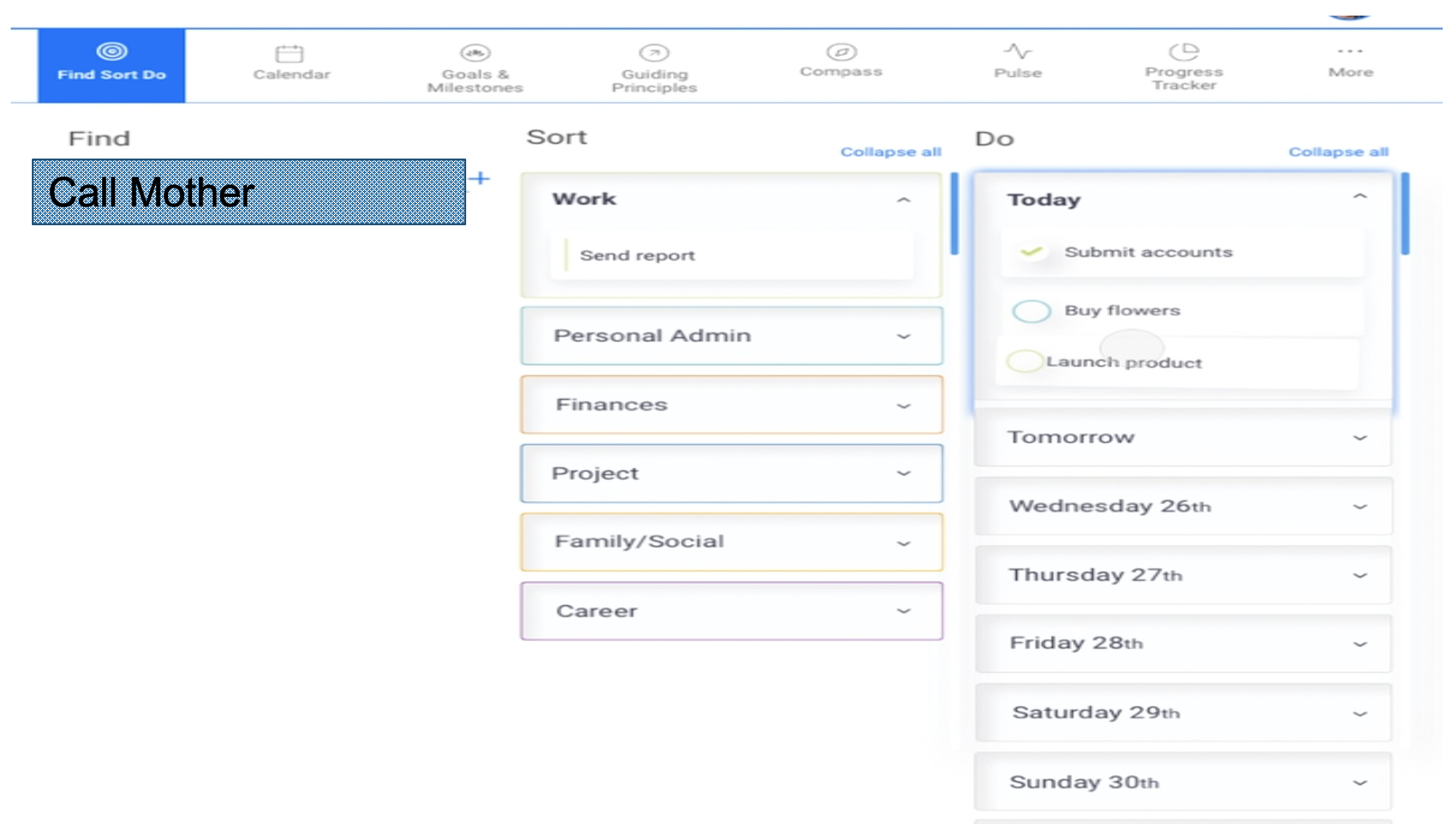}
     \caption{MGL illustration}
         \label{fig:nneg}
    \end{subfigure}
    \caption{Meta-Goal Learner.}
    \label{MGL}
\end{figure}

\begin{enumerate}
 
    \item  Tokenise the sentence and generate clauses as part of the background knowledge:\\ e.g.,
        contains(X,call).
        contains(X,mum).
    \item Send a request to ConceptNet web API for fetching the relations of the words separately and adding them to the background knowledge:\\ e.g.,
        related\_to(call, phone).
        related\_to(mother, family).
        
\end{enumerate}

Given the positive example and background knowledge, the system will learn the hypothesis using ILP. When the user adds the next sentence, the system will follow steps one and two and checks with the hypotheses learned before. If there is a match, the system will label the sentence as before; otherwise, the system will try to learn a new classification rule using the current input task as a positive example and examples from other tasks as negative examples. In the next section, we present some of the experimental evaluations of the learning algorithms which we have been exploring for integration within MGL. 

\subsection{ILP Experiments using the learning engine Aleph}

In this experiment, we check whether it is possible to use the ILP learning engine Aleph\cite{srinivasan2007aleph} for few-shot text classification. Examples and background knowledge are defined as follows:\\

\textbf{Examples} 
We use two positive examples for each of the "family", "work", and "sport" categories. For negative examples, we use every example of each category as the negative example for the other category.\\

\textbf{Background Knowledge}
To provide Background knowledge, we initially omit stop-words using NLTK library. Then, the system generates predicates for each sentence in two steps: 
\begin{enumerate}
    \item Indicating every word of the sentences in a "contains" predicate \\
    (e.g.  contains([registering,gym],gym)).
    \item Sending a request to ConceptNet API for each word of the sentence to fetch their "related" terms and then generating their "related\_to" predicate (e.g. related\_to(mother,family)).\\
\end{enumerate}

\textbf{Results and discussion} Using the learning engine Aleph, three rules (hypotheses) have been generated to classify the three categories.\\

\noindent Rule 1: category(A,family) :-   contains(A,B), related\_to(B,shop).\\
Rule 2: category(A,work) :-   contains(A,B), related\_to(B,letter).\\
Rule 3: category(A,sport) :-  contains(A,B), related\_to(B,exercise).\\

As explained above the training data for each category included 2 positive examples. We also experimented with one positive example (one-shot learning) but Aleph was unable to learn any rule. To evaluate the accuracy of the above rules on separate test data, we use a different dataset with 29 positive examples, including 15 "family", 9 "work" and 5 "sport" examples and the negative test data for each group consists of examples from other groups (i.e., 15 family examples used as negative for work, 9 work examples are used as negative for sport, and 5 sport examples are used as negative for family) and reached to the overall predictive accuracy of 64\%. 
\subsection{MIL Experiments using the learning engine Metagol}

In this experiment, we first check whether it is possible to learn a hypothesis from one positive and one negative example using MIL then explore the possibility of learning more general rules using predicate invention and recursive rules in MIL.\\

We employ the learning engine Metagol \cite{Pahlavi2014} to utilise MIL with the same Background Knowledge and examples applied in the Aleph experiment.
In addition to training examples and background knowledge, Metagol also needs metarules to induce the hypothesis. We test our system with two different types of metarules. At first, we utilise a \textit{chain} metarule to obtain the learning hypothesis. In the second experiment, we use \textit{indent}, \textit{chain}, and \textit{recursive} metarules along with using constants.\\

\textbf{Results and discussion}
The system induces one hypothesis from one positive and one negative example. Note that in this experiment, the positive example is a simple case where the category name is directly related to one of the words in the sentence. Consequently, one chain metarule is enough to generate the hypothesis. For instance, in category([call, mother], family), 'mother' directly connects to 'family' with only one edge of 'related\_to' relation in ConceptNet.  Hence, the acquired hypothesis consists of one 'related\_to' predicate as follows:\\

\noindent category(A,B) :- contains(A,C), related\_to(C,B).\\

This rule is similar to the rule which was learned by Aleph in the previous section when we provided at least two positive examples for each category and used variable in mode declaration but Aleph (unlike Metagol) was unable to learn from just one positive example. However, the rule above cannot cover all relevant positive examples as it only considers 'related\_to' once. 
To demonstrate this, we provided more training examples (4 positive and 6 negative) which include cases where we need at least two levels of connections, along with using chain metarule. 
Consequently, the system invents a new predicate (i.e., category\_1(A,B):- related\_to(A,C),related\_to(C,B).) within its hypothesis set as follows:\\

\noindent category(A,B) :- contains(A,C), category\_1(C,B).\\
category\_1(A,B):- related\_to(A,C),related\_to(C,B).
\paragraph{}The new rule can capture cases such as the category name 'home' which does not directly connect to 'mother' but is connected via 'family'. However, in order to have a general transitive rule for 'related\_to' we need to learn a recursive rule.\\

Finally, we demonstrate that Metagol can be used to learn a recursive rule to capture transitive 'related\_to' relations. In this experiment, we provide two positive examples with several relatednesses in their structure. Moreover, we use constant and variable at the indent metarule and consequently, the following results including a complete recursive hypothesis are achieved:\\

\noindent category(A,B):-contains(A,C),category\_1(C,B).\\
category\_1(A,B):-related\_to(A,C),category\_1(C,B).\\
category\_1(A,home):-related\_to(A,home).\\

As shown above, it is notable that whatever a chosen example for one-shot learning has more relatedness in its structure, i.e. higher complexity compared to the other examples, it could achieve a complete recursive hypothesis. This, in turn, is especially beneficial for choosing the best possible shot to do one-shot learning.\\

 To evaluate  one-shot learning, we employ the same test dataset used for the experiments with Aleph in the previous section. As evidenced by the results, Metagol could achieve around 70\% accuracy, while Aleph could not learn any rule from one example. \\  

In summary, we demonstrated that the Metagol learning engine can be used in our system to learn text classification rules from one example (one-shot learning). This feature clearly distinguishes Metagol (MIL) from Aleph (standard ILP). Moreover, we also demonstrated learning recursive rules in addition to predicate invention. However, this capability required metarules defined for the given application. In future, we will explore new implementations of MIL which can automatically generate the required metarules from background knowledge and examples. 
\subsection{Experiments comparing MIL with Deep Learning}

In this experiment, we use Siamese Network for few-shot learning for text classification and compare the results with MIL from the previous sections. A Siamese network includes two identical sub-networks with shared weights. Each sub-network works simultaneously and compares its outputs at the end \cite{Chicco2021}. If the input of the twin sub-networks is the same, they extract similar semantic features, and the distance between their output will be less; otherwise, the distance will be more significant. The Siamese Network used in this experiment is similar to the CNN-based Siamese model described in \cite{10.5555/3288251.3288303} where each twin sub-networks consist of a convolutional layer, a max pooling layer and a fully connected layer.
The model sets with a batch size of 50, and a learning rate of 0.00001. To achieve optimal results, the network is trained over ten epochs.\\

Following the approach described in \cite{varghese2021human}, we evaluate the performance of the Siamese network and compare the results with our one-shot learning approach using Metagol. The training data for both systems are one positive example and one or more negative examples (examples from other categories) and the test data is the same as the one used in the previous section.\\

\begin{figure}[h!]
    \centering
     \begin{subfigure}[b]{0.45\textwidth}
    \centering
    \includegraphics[width=\linewidth]{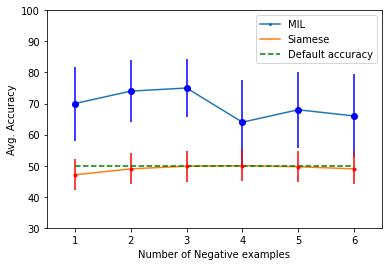}
     \caption{Increasing negative samples}
         \label{fig:newtask1}
    \end{subfigure}
    \begin{subfigure}[b]{0.45\textwidth}
    \centering
    \includegraphics[width=\linewidth]{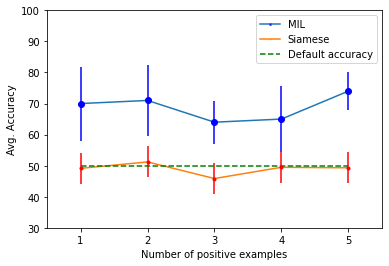}
    \caption{Increasing positive samples}
         \label{fig:newtask2}
    \end{subfigure}
    \caption{Comparison between the average predictive accuracy of MIL (Metagol) vs Deep Learning (Siamese Net) on Task Classification dataset. (a) learning with one positive example and one or more negative examples. (b) learning with one or more positive examples, and the same number of negatives. The training and test data have been randomly chosen from the Task Classification dataset and the test data includes 15 positive and 14 negative examples.}
    \label{newtask}
\end{figure}
As shown in Figure (\ref{fig:newtask1}), the Siamese model started from 47\% average accuracy with one positive and one negative example and stays around the default accuracy after learning from 6 examples(50\%). Our system achieved an average accuracy of 70\% from one positive and one negative example and maintained higher average accuracy than the Siamese model throughout. Figure (\ref{fig:newtask2}), depicts the average accuracy of our system and Siamese model in the case of an increasing number of positive examples. It also confirms the outperformance of our model over Siamese model.\\

\begin{figure}[h!]
    \centering
     \begin{subfigure}[b]{0.45\textwidth}
    \centering
    \includegraphics[width=\linewidth]{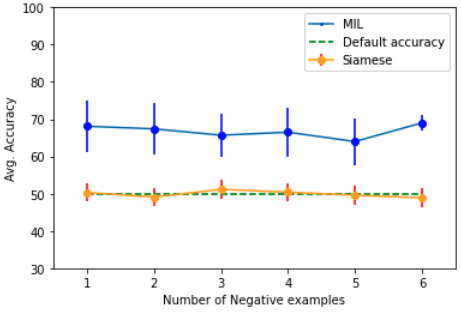}
     \caption{Increasing negative samples}
         \label{fig:nneg}
    \end{subfigure}
    \begin{subfigure}[b]{0.45\textwidth}
    \centering
    \includegraphics[width=\linewidth]{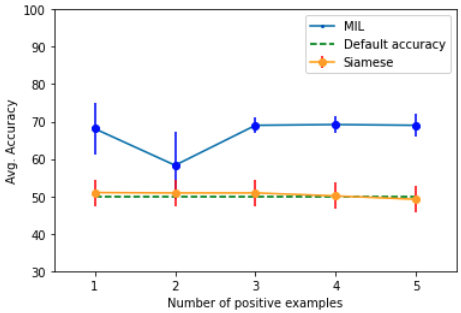}
    \caption{Increasing positive samples}
         \label{fig:nneg}
    \end{subfigure}
    \caption{Comparison between the average predictive accuracy of MIL (Metagol) vs Deep Learning (Siamese Net) on News Category dataset. (a) learning with one positive example and one or more negative examples. (b) learning with one or more positive examples, and the same number of negatives. The training and test data have been randomly chosen from News Category Dataset and the test data includes 50 positive and 50 negative examples.}
    \label{comparisonSiameseMILnews}
\end{figure}
To evaluate our model on a publicly available dataset, we use News Category Dataset \cite{misra2022news} which contains news articles from different sources and covers various topics such as sports, technology and the environment. We randomly selected five positive examples from the environment news headlines to train our model and test the gain hypotheses on randomly selected 50 positive and 50 negative samples. As shown in Figure \ref{comparisonSiameseMILnews}, the results suggest the advantage of our method based on MIL over the Siamese model when we have few training examples, e.g. one positive and one negative example.

It is noticeable that for training and evaluating a large dataset, we use two techniques as follows:
\begin{enumerate}
    \item To solve the problem of requiring large amounts of background knowledge for training and testing, we use the background knowledge of each example separately. 
    \item To train our model on a large dataset with only one or a few examples (one-shot learning), we randomly select 10 samples and make an average from several possible one-shots to calculate the average of different types of one-shots (ex: the average of possible one positive and one negative example shots or an average of one positive and two negative examples etc).

\end{enumerate}

\section{Conclusion}
Short text classification is challenging due to the freely constructed sentence structures and their limited length. This study aims to provide a novel algorithm to classify short texts called "tasks" automatically based on the user's interests and without having any prior information about user behaviours. Thus, the first research question sought to determine a reliable source for providing pertinent background information to feed the system. We utilised Conceptnet to prepare the required background information related to the task. Our model was then designed based on different types of ILP and MIL systems, including Aleph and Metagol. We also used a Siamese Network model to compare with the deep learning methods. Siamese nets are the most popular DL approaches for one or few-shot learning. We applied the Convolutional-Siamese model with one positive and one or more negative examples. 
According to the current results, the MIL-based approach achieved 70\% accuracy for one-shot learning, while the accuracy of ILP and Deep Learning was around default accuracy (50\%) in this task.

Finally, we used News Category Dataset as a publicly available dataset to evaluate our model. This led us to utilise two techniques, including Background Knowledge Splitting and the Average One Shot Learning approach to train and evaluate the large dataset. The final results validated MIL's superior performance to the Siamese network for one-shot learning from text.

\noindent\textbf{Supplemental Materials}\\
The data and code used in the experiments can be found in the following Github repository: \\\href{https://github.com/ghazalmilani/One-Shot-Learning-from-Text-ICLP2023.git}{https://github.com/ghazalmilani/One-Shot-Learning-from-Text-ICLP2023.git}

\noindent\textbf{Acknowledgement}\\
The first author would like to acknowledge EIT-Digital and Goalshaper Ltd, especially Ms Tracey Carr for supporting her PhD studies. We also thank Dany Varghese for the discussions related to this project.
\nocite{*}
\bibliographystyle{eptcs}
\bibliography{ref1}
\end{document}